\def\year{2020}\relax
\documentclass[letterpaper]{article} % DO NOT CHANGE THIS
\usepackage{aaai20}  % DO NOT CHANGE THIS
\usepackage{times}  % DO NOT CHANGE THIS
\usepackage{helvet} % DO NOT CHANGE THIS
\usepackage{courier}  % DO NOT CHANGE THIS
\usepackage[hyphens]{url}  % DO NOT CHANGE THIS
\usepackage{graphicx} % DO NOT CHANGE THIS
\usepackage{graphicx}  % DO NOT CHANGE THIS
\usepackage{bm,amsmath,amssymb}
\usepackage{xspace}
\usepackage{graphicx}
\usepackage{color}
\usepackage{algorithm}
\usepackage{algorithmic}

\usepackage{colortbl}
\definecolor{light_gray}{RGB}{170,170,170}

\usepackage[utf8]{inputenc} % allow utf-8 input
\usepackage{url}            % simple URL typesetting
\usepackage{booktabs}       % professional-quality tables
\usepackage{amsfonts}       % blackboard math symbols
\usepackage{nicefrac}       % compact symbols for 1/2, etc.
\usepackage{microtype}      % microtypography
\usepackage{longtable,lscape}
\usepackage{array,multirow}
\usepackage{txfonts}

\usepackage{subcaption}
\newcommand{\hide}[1]{}

\usepackage{enumitem}

\usepackage{bm}

\newcommand{\Ab}{\mathbf{A}}

\newcommand{\Cb}{\mathbf{C}}
\newcommand{\Db}{\mathbf{D}}

\newcommand{\Ib}{\mathbf{I}}

\newcommand{\Kb}{\mathbf{K}}

\newcommand{\Mb}{\mathbf{M}}

\newcommand{\Pb}{\mathbf{P}}
\newcommand{\Qb}{\mathbf{Q}}

\newcommand{\Vb}{\mathbf{V}}
\newcommand{\Wb}{\mathbf{W}}

\newcommand{\cb}{\mathbf{c}}

\newcommand{\vb}{\mathbf{v}}

\newcommand{\mname}{GCT\xspace}
\title{Learning the Graphical Structure of Electronic Health Records with\\Graph Convolutional Transformer}
\author{
Edward Choi$^1$, Zhen Xu$^1$, Yujia Li$^2$, Michael W. Dusenberry$^1$,\\
\Large \textbf{Gerardo Flores$^1$, Emily Xue$^1$, Andrew M. Dai$^1$}\\
$^1$ Google, Mountain View, USA $\quad$ $^2$ DeepMind, London, UK 
\vspace{-3mm}
}
\begin{document}

\maketitle

\begin{abstract}
%\vspace{-2mm}
Effective modeling of electronic health records (EHR) is rapidly becoming an important topic in both academia and industry.
A recent study showed that using the graphical structure underlying EHR data (\textit{e.g.} relationship between diagnoses and treatments) improves the performance of prediction tasks such as heart failure prediction.
However, EHR data do not always contain complete structure information. Moreover, when it comes to claims data, structure information is completely unavailable to begin with.
Under such circumstances, can we still do better than just treating EHR data as a flat-structured \textit{bag-of-features}?
In this paper, we study the possibility of jointly learning the hidden structure of EHR while performing supervised prediction tasks on EHR data.
Specifically, we discuss that Transformer is a suitable basis model to learn the hidden EHR structure, and propose Graph Convolutional Transformer, which uses data statistics to guide the structure learning process.
The proposed model consistently outperformed previous approaches empirically, on both synthetic data and publicly available EHR data, for various prediction tasks such as graph reconstruction and readmission prediction, indicating that it can serve as an effective general-purpose representation learning algorithm for EHR data.
\end{abstract}

\section{Introduction}
\label{sec:intro}
Large medical records collected by electronic healthcare records (EHR) systems in healthcare organizations enabled deep learning methods to show impressive performance in diverse tasks such as predicting diagnosis \cite{lipton2015learning,choi2016doctor,rajkomar2018scalable}, learning medical concept representations \cite{che2015deep,choi2016learning,choi2016multi,miotto2016deep}, and making interpretable predictions \cite{choi2016retain,ma2017dipole}.
As diverse as they are, one thing shared by all tasks is the fact that, under the hood, some form of neural network is processing EHR data to learn useful patterns from them.
To successfully perform any EHR-related task, it is essential to learn effective representations of various EHR features: diagnosis codes, lab values, encounters, and even patients themselves.
\begin{figure}[t]
\centering
\includegraphics[width=.47\textwidth]{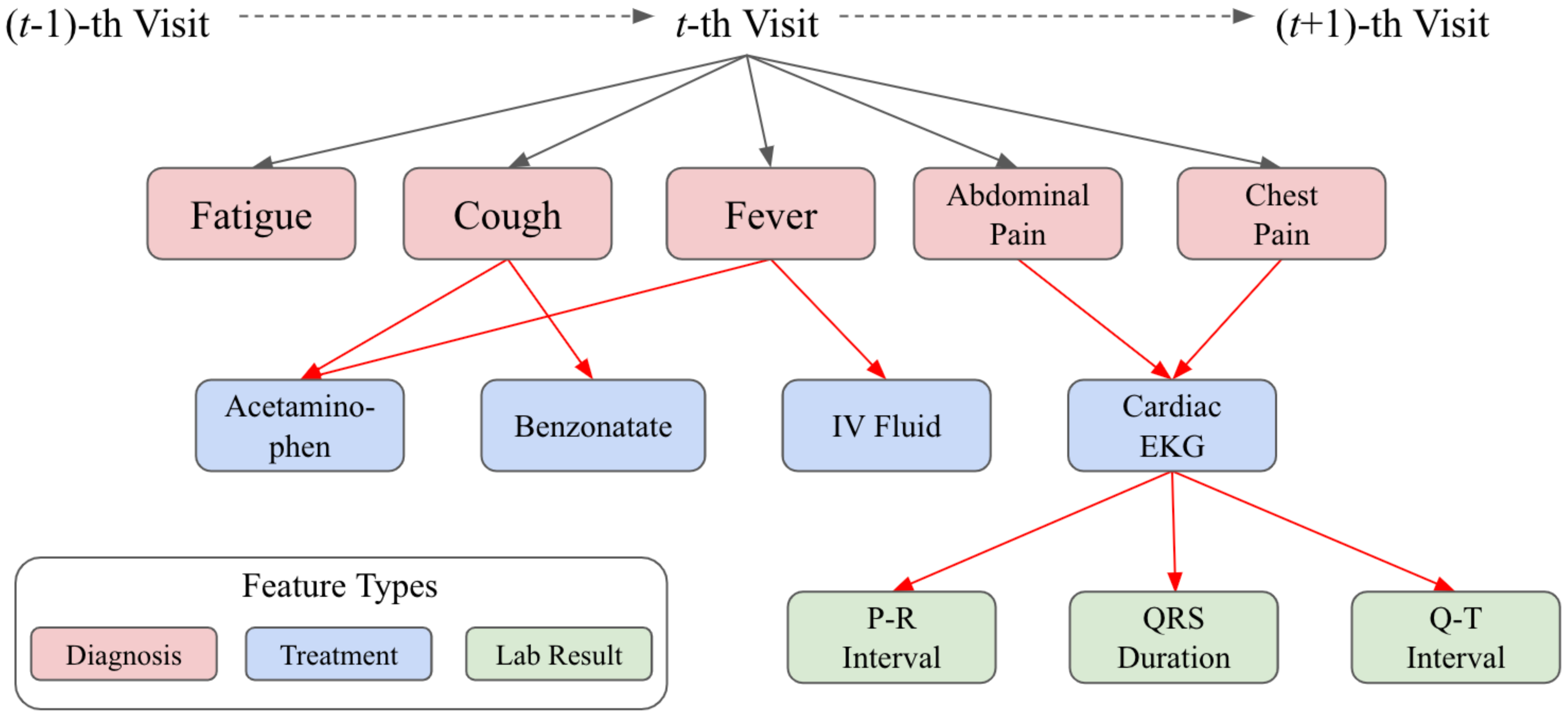}
\caption{
Graphical structure of electronic health records. A visit consists of multiple types of features, and their connections (red edges) reflect the physician's decision process.
}
\label{fig:ehr_graph}
%\vspace{-3mm}
\end{figure}
EHR data are typically stored in a relational database that can be represented as a hierarchical graph depicted in Figure~\ref{fig:ehr_graph}.
The common approach for processing EHR data with neural networks has been to treat each encounter as an unordered set of features, or in other words, a \textit{bag of features}.
However, the \textit{bag of features} approach completely disregards the graphical structure that reflects the physician's decision process.
For example, if we treat the encounter in Figure~\ref{fig:ehr_graph} as a \textit{bag of features}, we will lose the information that \textit{Benzonatate} was ordered because of \textit{Cough}, not because of \textit{Abdominal pain}.

Recently, motivated by this EHR structure, \cite{choi2018mime} proposed MiME, a model architecture that reflects EHR's encounter structure, specifically the relationships between the diagnosis and its treatment. MiME outperformed various \textit{bag of features} approaches in prediction tasks such as heart failure diagnosis prediction.
Their study, however, naturally raises the question: when the EHR data do \textit{not} contain structure information (the red edges in Figure~\ref{fig:ehr_graph}), can we still do better than \textit{bag of features} in learning the representation of the data for various prediction tasks?
This question emerges in many occasions, since EHR data do not always contain the entire structure information. For example, some dataset might describe which treatment lead to measuring certain lab values, but might not describe the reason diagnosis for ordering that treatment.
Moreover, when it comes to claims data, such structure information is completely unavailable to begin with.

To address this question, we propose Graph Convolutional Transformer (\mname), a novel approach to jointly learn the hidden encounter structure while performing various prediction tasks when the structure information is unavailable.
Throughout the paper, we make the following contributions:
\begin{itemize}
\item To the best of our knowledge, this is the first work to successfully perform the joint learning of the hidden structure of EHR data and a supervised prediction task.
\item We propose a novel modification to the Transformer to guide the self-attention to learn the hidden EHR structure using prior knowledge in the form of attention masks and prior conditional probability based regularization. And we empirically show that unguided self-attention alone cannot properly learn the hidden EHR structure.
\item \mname outperforms baseline models in all prediction tasks (\textit{e.g.} graph reconstruction, readmission prediction) for both the synthetic dataset and a publicly available EHR dataset, showing its potential to serve as an effective general-purpose representation learning algorithm for EHR data.
\end{itemize}
%After we discuss related works in the next section, we describe the graphical nature of encounter records, and revisit the connection between Transformer's self-attention \cite{vaswani2017attention} and Graph Convolutional Networks (GCN) \cite{kipf2016semi} to discuss learning of the hidden encounter structure.
%fThen we describe \mname to more effectively utilize the characteristics of EHR data while performing diverse prediction tasks.

\section{Related Work}
\label{sec:related}
Although there are works on medical concept embedding, focusing on patients \cite{che2015deep,miotto2016deep,suresh2017clinical,nguyen2018resset}, visits \cite{choi2016multi}, or codes \cite{tran2015learning,choi2017gram,shang2019gamenet}, the graphical nature of EHR has not been fully explored yet.
\cite{choi2018mime} proposed MiME, which derives the visit representation in a bottom-up fashion according to the encounter structure.
For example in Figure~\ref{fig:ehr_graph}, MiME first combines the embedding vectors of lab results with the \textit{Cardiac EKG} embedding, which is then combined with both \textit{Abdominal Pain} embedding and \textit{Chest Pain} embedding.
Then all diagnosis embeddings are pooled together to derive the final visit embedding.
By outperforming various bag-of-features models in various prediction tasks, MiME demonstrated the importance of the structure information of encounter records.

\textit{Transformer} \cite{vaswani2017attention} was proposed for machine translation.
It uses a novel method to process sequence data using only attention \cite{bahdanau2014neural}, and recently showed impressive performance in other tasks such as BERT (\textit{i.e.} pre-training word representations) \cite{devlin2018bert}.
There are recent works that use Transformer on medical records \cite{song2018attend,wang2019inpatient2vec,shang2019pre,li2019behrt}, but they either simply replace RNN with Transformer to handle ICU records, or directly apply BERT learning objective on medical records, and do not utilize the hidden structure of EHR.
\textit{Graph (convolutional) networks} encompass various neural network methods to handle graphs such as molecular structures, social networks, or physical experiments. \cite{kipf2016semi,hamilton2017inductive,battaglia2018relational,xu2018powerful}. 
In essence, many graph networks can be described as different ways to aggregate a given node's neighbor information, combine it with the given node, and derive the node's latent representation \cite{xu2018powerful}.

Some recent works focused on the connection between the Transformer's self-attention and graph networks \cite{battaglia2018relational}.
Graph Attention Networks \cite{velivckovic2017graph} applied self-attention on top of the adjacency matrix to learn non-static edge weights, and \cite{wang2018non} used self-attention to capture non-local dependencies in images.
Although our work also makes use of self-attention, \mname's objective is to jointly learn the underlying structure of EHR even when the structure information is missing, and better perform supervised prediction tasks, and ultimately serve as a general-purpose EHR embedding algorithm.
In the next section, we outline the graphical nature of EHR, then revisit the connection between Transformer and GCN to motivate the EHR structure learning, after which we describe \mname.

\section{Method}
\label{sec:method}
\subsection{Electronic Health Records as a Graph}
%Electronic Health Records (EHR) data are typically stored in a relational database, which consists of multiple tables such as encounter, medication orders, procedure orders and lab results. The tables can be linked with foreign keys (\textit{e.g.} encounter ID, medication order ID) to derive a graphical structure such as Figure~\ref{fig:ehr_graph}. 
As depicted in Figure~\ref{fig:ehr_graph}, the $t$-th visit $\mathcal{V}^{(t)}$ starts with the visit node $v^{(t)}$ at the top. Beneath the visit node are diagnosis nodes $d^{(t)}_1, \ldots, d^{(t)}_{|d^{(t)}|}$, which lead to ordering a set of treatments $m^{(t)}_1, \ldots, m^{(t)}_{|m^{(t)}|}$, where $|d^{(t)}|, |m^{(t)}|$ respectively denote the number of diagnosis and treatment codes in $\mathcal{V}^{(t)}$.
Some treatments produce lab results $r^{(t)}_1, \ldots, r^{(t)}_{|r^{(t)}|}$, which may be associated with continuous values (\textit{e.g.} blood pressure) or binary values (\textit{e.g.} positive/negative allergic reaction). 
Since we focus on a single encounter in this study
\footnote{Note that a sequence aggregator such as RNN or 1-D CNN can convert a sequence of individual encounters $\mathcal{V}^{(0)}, \ldots \mathcal{V}^{(t)}$ to a patient-level representation.}
, we omit the time index $t$ throughout the paper.

If we assume all features $d_i$, $m_i$, $r_i$\footnote{If we bucketize the continuous values of $r_i$, we can treat $r_i$ as a discrete feature like $d_i$, $m_i$.} can be represented in the same latent space, then we can view an encounter as a graph consisting of $|d| + |m| + |r|$ nodes with an adjacency matrix $\Ab$ that describes the connections between the nodes. We use $c_i$ as the collective term to refer to any of $d_i$, $m_i$, and $r_i$ for the rest of the paper.
Given $c_i$ and $\Ab$, we can use graph networks or MiME to derive the visit representation $\vb$ and use it for downstream tasks such as heart failure prediction.
However, if we do not have the structural information $\Ab$, which is the case in many EHR data and claims data, we typically use feed-forward networks to derive $\vb$, which is essentially summing all node representations $\cb_i$'s and projecting it to some latent space.
\begin{figure}[t]
\centering
\includegraphics[width=.47\textwidth]{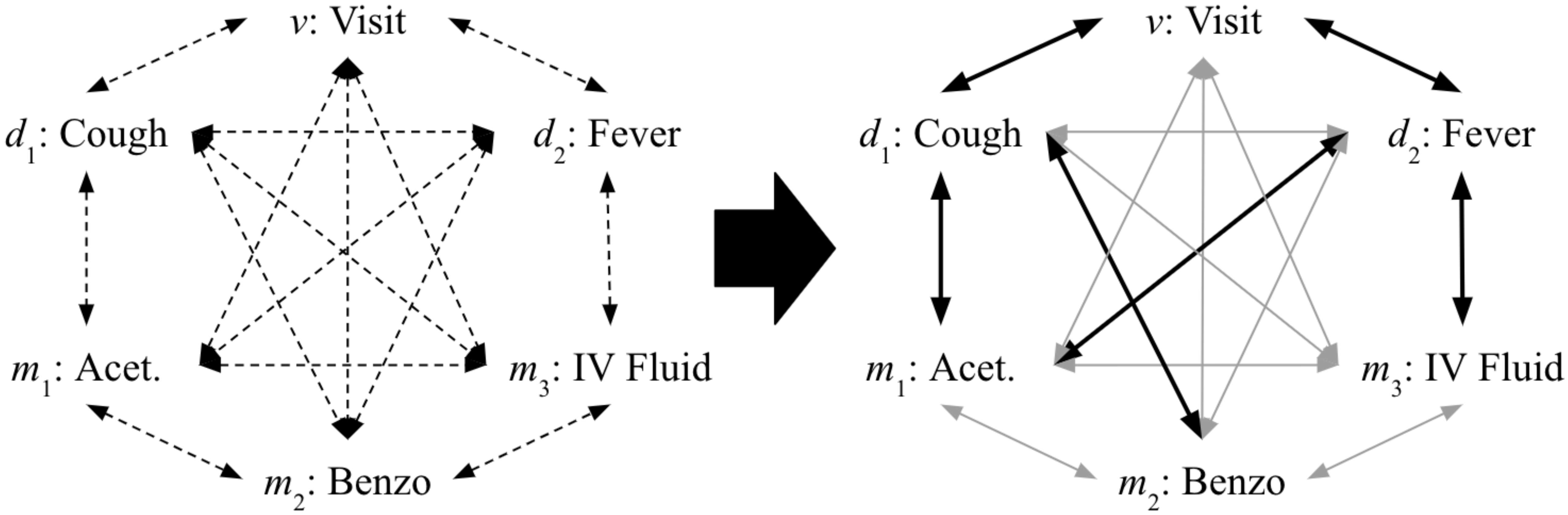}
\caption{
Learning the underlying structure of an encounter. Self-attention can be used to start from the left, where all nodes are fully-connected, and arrive at the right, where meaningful connections are in thicker edges.
}
\label{fig:fully_connected}
%\vspace{-3mm}
\end{figure}
\subsection{Transformer and Graph Networks}
Even without the structure information $\Ab$, it is unreasonable to treat $\mathcal{V}$ as a bag of nodes $c_i$, because obviously physicians must have made some decisions when making diagnosis and ordering treatments. The question is how to utilize the underlying structure without explicit $\Ab$.
One way to view this problem is to assume that all nodes $c_i$ in $\mathcal{V}$ are implicitly fully-connected, and try to figure out which connections are stronger than the other as depicted in Figure~\ref{fig:fully_connected}.
In this work, as discussed in section~\ref{sec:related}, we use Transformer to learn the underlying encounter structure.
To elaborate, we draw a comparison between two cases:
\begin{itemize}[leftmargin=5.5mm]
\item Case A: We know $\Ab$, hence we can use Graph Convolutional Networks (GCN). In this work, we use multiple hidden layers between each convolution, motivated by \cite{xu2018powerful}.
\begin{equation}
\label{eq:gin}
\Cb^{(j)} = \mbox{MLP}^{(j)}(\tilde{\Db}^{-1} \tilde{\Ab} \Cb^{(j-1)} \Wb^{(j)}),
%\Cb^{(j)} = \mbox{MLP}^{(j)}(\hat{\Db}^{-1} \hat{\Ab} \Cb^{(j-1)}),
\end{equation}
where $\tilde{\Ab} = \Ab + \Ib$, $\tilde{\Db}$ is the diagonal node degree matrix\footnote{\cite{xu2018powerful} does not use the normalizer $\tilde{\Db}^{-1}$ to improve model expressiveness on multi-set graphs, but we include $\tilde{\Db}^{-1}$ to make the comparison with Transformer clearer.} of $\tilde{\Ab}$, $\Cb^{(j)}$ and $\Wb^{(j)}$ are the node embeddings and the trainable parameters of the $j$-th convolution respectively.
MLP$^{(j)}$ is a multi-layer perceptron of the $j$-th convolution with its own trainable parameters.
\item Case B: We do not know $\Ab$, hence we use Transformer, specifically the encoder with a single-head attention, which can be formulated as
\begin{equation}
\label{eq:transformer}
\Cb^{(j)} = \mbox{MLP}^{(j)}(\operatorname{softmax}(\frac{\Qb^{(j)} \Kb^{(j)\top}}{\sqrt{d}}) \Vb^{(j)}),
%\Cb^{(j)} = \mbox{MLP}^{(j)}(\operatorname{softmax}(\frac{\Qb^{(j)} \Kb^{(j)\top}}{\sqrt{d}})),
\end{equation}
where $\Qb^{(j)} = \Cb^{(j-1)} \Wb^{(j)}_{Q}$, $\Kb^{(j)} = \Cb^{(j-1)} \Wb^{(j)}_{K}$, $\Vb^{(j)} = \Cb^{(j-1)} \Wb^{(j)}_{V}$, and $d$ is the column size of $\Wb^{(j)}_{K}$.
$\Wb^{(j)}_{Q}, \Wb^{(j)}_{K}$, and $\Wb^{(j)}_{V}$ are trainable parameters of the $j$-th Transformer block\footnote{Since we use MLP in both GCN and Transformer, the terms $\Wb^{(j)}$ and $\Wb^{(j)}_{V}$ are unnecessary, but we put them to follow the original formulations.}.
Note that positional encoding using sine and cosine functions is not required, since features in an encounter are unordered.
\end{itemize}
Given Eq.~\ref{eq:gin} and Eq.~\ref{eq:transformer}, we can readily see that there is a correspondence between the normalized adjacency matrix $\tilde{\Db}^{-1} \tilde{\Ab}$ and the attention map $\operatorname{softmax}(\frac{\Qb^{(j)} \Kb^{(j)T}}{\sqrt{d}})$,
and between the node embeddings $\Cb^{(j-1)} \Wb^{(j)}$ and the value vectors $\Cb^{(j-1)} \Wb^{(j)}_{V}$.
Therefore GCN can be seen as a special case of Transformer, where the attention mechanism is replaced with the known, fixed adjacency matrix.
Conversely, Transformer can be seen as a graph embedding algorithm that assumes fully-connected nodes and learns the connection strengths during training \cite{battaglia2018relational}.
Given this connection, it seems natural to take advantage of Transformer as a base algorithm to learn the underlying structure of visits.

%Note that Transformer's self-attention has been used in previous works for learning relations between features in settings other than text. 
%Graph Attention Networks \cite{vaswani2017attention} applied attention on top of the adjacency matrix to learn non-static edge weights, and \cite{wang2018non} used self-attention to capture non-local dependencies in images.
%Although our work also relies on attention, our interest lies in whether Transformer can be an effective tool to capture the underlying graphical structure of EHR data even when the structural information is missing, thus improving encounter-based prediction tasks.
\begin{figure*}[t]
\centering
\includegraphics[width=.7\textwidth]{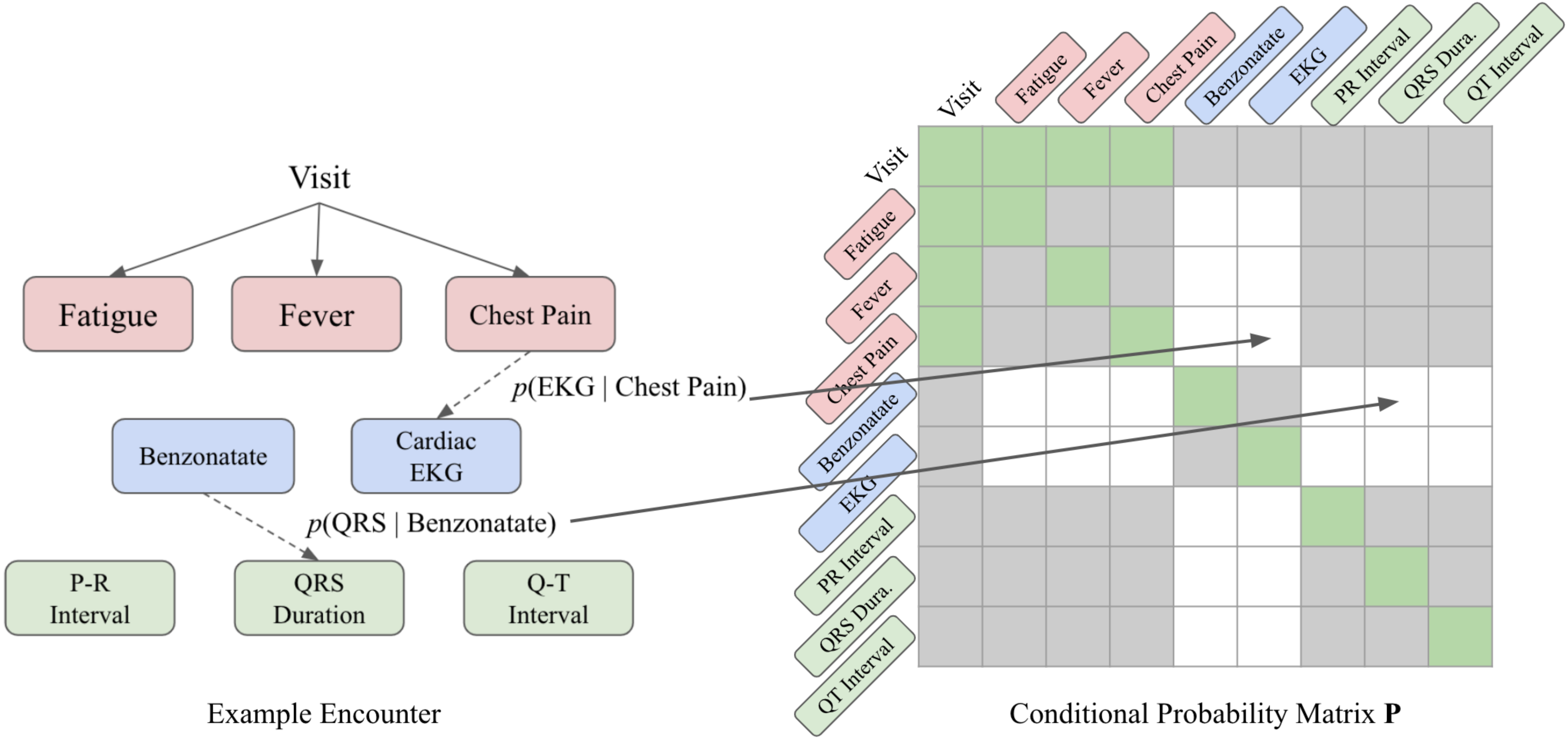}
\caption{
Creating the conditional probability matrix $\Pb$ based on an example encounter. The gray cells are masked to zero probability since those connections are not allowed. The green cells are special connections that we know are guaranteed to exist. We assign a pre-defined scalar value (\textit{e.g.} 1) to the green cells. The white cells are assigned the corresponding conditional probabilities.
}
\label{fig:conditional_matrix}
\end{figure*}
\subsection{Graph Convolutional Transformer}
Although Transformer can potentially learn the hidden encounter structure, without a single piece of hint, it must search the entire attention space to discover meaningful connections between encounter features.
Therefore we propose Graph Convolutional Transformer (\mname), which, based on data statistics, restricts the search to the space where it is likely to contain meaningful attention distribution.

Specifically, we use 1) the characteristic of EHR data and 2) the conditional probabilities between features.
First, we use the fact that some connections are not allowed in the encounter record.
For example, we know that treatment codes can only be connected to diagnosis codes, but not to other treatment codes.
Based on this observation, we can create a mask $\Mb$, which will be used during the attention generation step.
$\Mb$ has negative infinities where connections are not allowed, and zeros where connections are allowed.

Conditional probabilities can be useful for determining potential connections between features.
For example, given \textit{chest pain}, \textit{fever} and \textit{EKG}, without any structure information, we do not know which diagnosis is the reason for ordering \textit{EKG}.
However, we can calculate from EHR data that $p(\mbox{EKG} | \mbox{chest pain})$ is typically larger than $p(\mbox{EKG} | \mbox{fever})$, indicating that the connection between the former pair is more likely than the latter pair.
Therefore we propose to use the conditional probabilities calculated from the encounter records as the guidance for deriving the attention.
%If we denote $p(m_i | d_j)$ as the probability of an $i$-th treatment given $j$-th diagnosis, then the conditional probability for all treatments and diagnoses, $p(m | d) \in [0.0, 1.0]^{|d| \times |m|}$ is a matrix.
%In the same fashion, we can calculate $p(d | m)$, $p(r | m)$, and $p(m | r)$.
%fNote that we assign zeros to disallowed connections, for example, $p(d | d)$ or $p(m | m)$, to have the same effect as applying $\Mb$.
After calculating $p(m | d), p(d | m), p(r | m)$, and $p(m | r)$ from all encounter records for all diagnosis codes $d$, treatment codes $m$, and lab codes $r$, we can create a guiding matrix when given an encounter record, as depicted by Figure~\ref{fig:conditional_matrix}.
We use $\Pb \in [0.0, 1.0]^{|c| \times |c|}$ to denote the matrix of conditional probabilities of all features, normalized such that each row sums to $1$.
Note that \mname's attention $\mbox{softmax}(\frac{\Qb^{(j)} \Kb^{(j) \top}}{\sqrt{d}})$, the mask $\Mb$, and the conditional probabilities $\Pb$ are of the same size.

Given $\Mb$ and $\Pb$, we want to guide \mname to recover the true graph structure as much as possible.
But we also want to allow some room for \mname to learn novel connections that are helpful for solving given prediction tasks.
Therefore \mname uses the following formulation:
%\begin{equation}
%\label{eq:define_adj}
%\mbox{Define } \hat{\Ab}^{(j)} := \mbox{softmax}(\frac{\Qb^{(j)} \Kb^{(j) \top}}{\sqrt{d}} + \Mb)
%\end{equation}
\begin{align}
%\label{eq:gct}
\mbox{Define } \hat{\Ab}^{(j)} & := \mbox{softmax}(\frac{\Qb^{(j)} \Kb^{(j) \top}}{\sqrt{d}} + \Mb) \label{eq:define_adj} \\
\mbox{Self-attention:} \nonumber\\
\Cb^{(j)} &= \mbox{MLP}^{(j)} \bigg( \Pb \Cb^{(j-1)} \Wb_{V}^{(j)} \bigg) \mbox{ when $j = 1$}, \nonumber\\
\Cb^{(j)} &= \mbox{MLP}^{(j)} \bigg( \hat{\Ab}^{(j)} \Cb^{(j-1)} \Wb_{V}^{(j)} \bigg) \mbox{ when $j > 1$} \nonumber \\
%\Cb^{(j)} &= \mbox{MLP}^{(j)} \bigg( \Pb \Cb^{(j-1)} \Wb_{V}^{(j-1)} \bigg) \mbox{, when $j = 1$} \nonumber \\
%f\Cb^{(j)} &= \mbox{MLP}^{(j)} \bigg( \hat{\Ab}^{(j)} \Cb^{(j-1)} \Wb_{V}^{(j)} \bigg) \mbox{, when $j > 1$} \nonumber \\
\mbox{Regularization:} \nonumber\\
L_{reg}^{(j)} &= D_{KL}(\Pb || \hat{\Ab}^{(j)}) \mbox{ when $j = 1$}, \nonumber\\
L_{reg}^{(j)} &= D_{KL}(\hat{\Ab}^{(j-1)} || \hat{\Ab}^{(j)}) \mbox{ when $j > 1$} \nonumber \\
%L_{reg}^{(j)} &= D_{KL}(\Pb || \hat{\Ab}^{(j)}) \mbox{, when $j = 1$} \nonumber\\
%L_{reg}^{(j)} &= D_{KL}(\hat{\Ab}^{(j-1)} || \hat{\Ab}^{(j)}) \mbox{, when $j > 1$}\nonumber\\
L &= L_{pred} + \lambda \sum_{j} L_{reg}^{(j)} \label{eq:gct}
\end{align}
In preliminary experiments, we noticed that attentions were often uniformly distributed in the first block of Transformer.
This seemed due to Transformer not knowing which connections were worth attending.
Therefore we replace the attention mechanism in the first \mname block with the conditional probabilities $\Pb$.
The following blocks use the masked self-attention mechanism.
However, we do not want \mname to drastically deviate from the informative $\Pb$, but rather gradually improve upon $\Pb$.
Therefore, based on the fact that attention is itself a probability distribution, and inspired by Trust Region Policy Optimization \cite{schulman2015trust}, we sequentially penalize attention of $j$-th block if it deviates too much from the attention of $j-1$-th block, using KL divergence.
As shown by Eq.~\eqref{eq:gct}, the regularization terms are summed to the prediction loss term (\textit{e.g.} negative log-likelihood), and the trade-off is controlled by the coefficient $\lambda$. 
\mname's code and datasets are publicly available at \url{https://github.com/Google-Health/records-research}.

\section{Experiments}
\label{sec:experiments}
\subsection{Synthetic Encounter Record}
\label{ssec:exp_synthetic}
MiME \cite{choi2018mime} was evaluated on proprietary EHR data that contained structure information.
Unfortunately, to the best of our knowledge, there are no public EHR data that contain structure information (which is the main motivation of this work).
To evaluate \mname's ability to learn EHR structure, we instead generated synthetic data that has a similar structure as EHR data.

The synthetic data has the same \textit{visit-diagnosis-treatment-lab results} hierarchy as EHR data, and was generated in a top-down fashion.
Each level was generated conditioned on the previous level, where the probabilities were modeled with the Pareto distribution, which
follows the power law that best captures the long-tailed nature of medical codes.
Using 1,000 diagnosis, treatment and lab codes each, we initialized $p(D), p(D|D), p(M|D), p(R|M,D)$ to follow the Pareto distribution, where $D, M$, and $R$ respectively denote diagnosis, treatment, and lab random variables.
$p(D)$ is used to draw independent diagnosis codes $d_i$, and $p(D|D)$ is used to draw $d_j$ that are likely to co-occur with the previously sampled $d_i$.
$P(M|D)$ is used to draw a treatment code $m_j$, given some $d_i$.
$P(R|M,D)$ is used to draw a lab code $r_k$, given some $m_j$ and $d_i$.
A further description of synthetic records is provided in Appendix~\ref{appendix:synthetic}.
Table~\ref{table:data_stat} summarizes the data statistics.

\subsection{eICU Collaborative Research Dataset}
\label{ssec:exp_eicu}
\begin{table}[t]
\caption{Statistics of the synthetic dataset and eICU}
\label{table:data_stat}
\footnotesize
\centering
\begin{tabular}{lcc}
% \hline
\toprule
& Synthetic & eICU\\
% \hline
\midrule
\# of encounters & 50,000 & 41,026\\
\# of diagnosis codes & 1,000 & 3,093 \\
\# of treatment codes & 1,000 & 2,132 \\
\# of lab codes & 1,000 & N/A \\
% \hline
\midrule
Avg. \# of diagnosis per visit & 7.93 & 7.70 \\
Avg. \# of treatment per visit & 14.59 & 5.03 \\
Avg. \# of lab per visit & 21.31 & N/A \\
% \hline
\bottomrule
\end{tabular}
\end{table}
\begin{table*}[t]
\caption{Graph reconstruction and diagnosis-treatment classification performance. Parentheses denote standard deviations. We report the performance measured in AUROC in Appendix~\ref{appendix:auroc}.}
\label{table:graph_recon_dt_pred_result}
\footnotesize
%\resizebox{\columnwidth}{!}{%
\centering
\begin{tabular}{lcccc}
% \hline
\toprule
& \multicolumn{2}{c}{\textbf{Graph reconstruction}} & \multicolumn{2}{c}{\textbf{Diagnosis-Treatment classification}}\\
%\hline
Model & Validation AUCPR & Test AUCPR & Validation AUCPR & Test AUCPR\\
%Model & \begin{tabular}{@{}c@{}} Validation\\AUCPR\end{tabular} & \begin{tabular}{@{}c@{}} Test\\AUCPR\end{tabular} & \begin{tabular}{@{}c@{}} Validation\\AUCPR\end{tabular} & \begin{tabular}{@{}c@{}} Test\\AUCPR\end{tabular}\\
% \hline
\midrule
GCN & 1.0 (0.0) & 1.0 (0.0) & 1.0 (0.0) & 1.0 (0.0)\\
% \hline
\midrule
GCN$_P$ & 0.5807 (0.0019) & 0.5800 (0.0021) & 0.8439 (0.0166) & 0.8443 (0.0214)\\
GCN$_{random}$ & 0.5644 (0.0018) & 0.5635 (0.0021) & 0.7839 (0.0144) & 0.7804 (0.0214)\\
Shallow & 0.5443 (0.0015) & 0.5441 (0.0017) & 0.8530 (0.0181) & 0.8555 (0.0206)\\
Deep & - & - & 0.8210 (0.0096) & 0.8198 (0.0046)\\
Transformer & 0.5755 (0.0020) & 0.5752 (0.0015) & 0.8329 (0.0282) & 0.8380 (0.0178)\\
\mname & \textbf{0.5972} (0.0027) & \textbf{0.5965} (0.0031) & \textbf{0.8686} (0.0103) & \textbf{0.8671} (0.0247)\\
% \hline
\bottomrule
\end{tabular}
%f}
\end{table*}
To test \mname on real-world EHR records, we use Philips eICU Collaborative Research Dataset\footnote{https://eicu-crd.mit.edu/about/eicu/} \cite{pollard2018eicu}.
eICU consists of Intensive Care Unit (ICU) records collected from multiple sites in the United States between 2014 and 2015.
From the encounter records, medication orders and procedure orders, we extracted diagnosis codes and treatment codes (\textit{i.e.} medication, procedure codes).
Since the data were collected from an ICU, a single encounter can last several days, where the encounter structure evolves over time, rather than being fixed as Figure~\ref{fig:ehr_graph}.
Therefore we used encounters that lasted less than 24 hours, and removed duplicate codes (\textit{i.e.} medications administered multiple times).
Additionally, we did not use lab results as their values change over time in the ICU (\textit{i.e.} blood pH level).
We leave as future work how to handle ICU records that evolve over a longer period of time.
Note that eICU does not contain structure information, \textit{e.g.} we know \textit{cough} and \textit{acetaminophen} in Figure~\ref{fig:ehr_graph} occur in the same visit, but do not know if \textit{acetaminophen} was prescribed due to \textit{cough}.
Table~\ref{table:data_stat} summarizes the data statistics.

\subsection{Baseline Models}
\label{ssec:exp_baselines}
\begin{itemize}[leftmargin=5.5mm]
\item \textbf{GCN}: Given the adjacency matrix $\Ab$, we follow Eq.~\eqref{eq:gin} to learn the feature representations $\cb_i$ of each feature $c_i$ in a visit $\mathcal{V}$.
The visit embedding $\vb$ (\textit{i.e.} graph-level representation) is obtained from the placeholder visit node $v$.
This will serve as the optimal model during the experiments. Note that MiME can be seen as a special case of GCN using unidirectional edges (\textit{i.e.} triangular adjacency matrix), and a special function to fuse diagnosis and treatment embeddings.
\item \textbf{GCN$_{P}$}: Instead of the true adjacency matrix $\Ab$, we use the conditional probability matrix $\Pb$, and follow Eq.~\eqref{eq:gin}. This will serve as a model that only relies on data statistics without any attention mechanism, which is the opposite of Transformer.
\item \textbf{GCN$_{random}$}: Instead of the true adjacency matrix $\Ab$, we use a randomly generated normalized adjacency matrix where each element is indepdently sampled from a uniform distribution between 0 and 1.
This model will let us evaluate whether true encounter structure is useful at all.
\item \textbf{Shallow}: Each $c_i$ is converted to a latent representation $\cb_i$ using multi-layer feedforward networks with ReLU activations.
The visit representation $\vb$ is obtained by simply summing all $\cb_i$'s.
We use layer normalization \cite{ba2016layer}, drop-out \cite{srivastava2014dropout} and residual connections \cite{he2016deep} between layers.
%This is equivalent to taking the node embeddings $\cb_i$'s (\textit{i.e.} columns of $\Wb_x$) in the visit $\mathcal{V}$, and summing them up to derive a single vector $\vb$.
%\item \textbf{non-linear}: The binary vector $\xb$ is transformed to a visit embedding $\vb = \sigma(\Wb_x \xb)$ where we use \textit{ReLU} for $\sigma(\cdot)$ to add non-linearity to \textbf{linear}.
\item \textbf{Deep}: We use multiple feedforward layers with ReLU activations (including layer normalization, drop-out and residual connections) on top of \textbf{shallow} to increase the expressivity.
Note that \cite{zaheer2017deep} theoretically describes that this model is sufficient to obtain the optimal representation of a set of items (\textit{i.e.,} a visit consisting of multiple features).
\end{itemize}

\begin{table*}[t]
\caption{Masked diagnosis code prediction performance on the two datasets. Parentheses denote standard deviations.}
\label{table:node_pred_result}
\footnotesize
%\resizebox{\columnwidth}{!}{%
\centering
\begin{tabular}{lcccc}
% \hline
\toprule
%Dataset & Model & \begin{tabular}{@{}c@{}} Validation\\Accuracy\end{tabular} & \begin{tabular}{@{}c@{}} Test\\Accuracy\end{tabular}\\

& \multicolumn{2}{c}{\textbf{Synthetic}} & \multicolumn{2}{c}{\textbf{eICU}}\\
%\hline
Model & Validation Accuracy & Test Accuracy & Validation Accuracy & Test Accuracy\\
%Model & \begin{tabular}{@{}c@{}} Validation\\Accuracy\end{tabular} & \begin{tabular}{@{}c@{}} Test\\Accuracy\end{tabular} & \begin{tabular}{@{}c@{}} Validation\\Accuracy\end{tabular} & \begin{tabular}{@{}c@{}} Test\\Accuracy\end{tabular}\\
% \hline
\midrule
GCN & 0.2862 (0.0048) & 0.2834 (0.0065) & - & -\\
% \hline
\midrule
GCN$_P$ & 0.2002 (0.0024) & 0.1954 (0.0064) & 0.7434 (0.0072) & 0.7432 (0.0086)\\
GCN$_{random}$ &  0.1868 (0.0031) & 0.1844 (0.0058) & 0.7129 (0.0044) & 0.7186 (0.0067)\\
Shallow & 0.2084 (0.0043) & 0.2032 (0.0068) & 0.7313 (0.0026) & 0.7364 (0.0017)\\
Deep & 0.1958 (0.0043) & 0.1938 (0.0038) & 0.7309 (0.0050) & 0.7344 (0.0043)\\
Transformer & 0.1969 (0.0045) & 0.1909 (0.0074) & 0.7190 (0.0040) & 0.7170 (0.0061)\\
\mname & \textbf{0.2220} (0.0033) & \textbf{0.2179} (0.0071) & \textbf{0.7704} (0.0047) & \textbf{0.7704} (0.0039)\\
% \hline
\bottomrule
 %& GCN & 0.2828 & 0.3024\\
 %f& GCN$_P$ & 0.1908 & 0.2096\\
 %f& shallow & 0.3488 & 0.3417\\
%feICU & deep & 0.3549 & 0.3547\\
 %f& Transformer & 0.3647 & 0.3581\\
 %f& \mname & 0.2182 & 0.2328\\
%f\hline
\end{tabular}
%f}
\end{table*}
\begin{table*}[t]
\caption{Readmission prediction and mortality prediction performance on eICU. Parentheses denote standard deviation. We report the performance measured in AUROC in Appendix~\ref{appendix:auroc}.}
\label{table:mortality_readmission_result}
\footnotesize
%\resizebox{\columnwidth}{!}{%
\centering
\begin{tabular}{lcccc}
% \hline
\toprule
& \multicolumn{2}{c}{\textbf{Readmission prediction}} & \multicolumn{2}{c}{\textbf{Mortality prediction}}\\
%\hline
Model & Validation AUCPR & Test AUCPR & Validation AUCPR & Test AUCPR\\
%Model & \begin{tabular}{@{}c@{}} Validation\\AUCPR\end{tabular} & \begin{tabular}{@{}c@{}} Test\\AUCPR\end{tabular} & \begin{tabular}{@{}c@{}} Validation\\AUCPR\end{tabular} & \begin{tabular}{@{}c@{}} Test\\AUCPR\end{tabular}\\
% \hline
\midrule
GCN$_P$ & 0.5121 (0.0154) & 0.4987 (0.0105) & 0.5808 (0.0331) & 0.5647 (0.0201)\\
GCN$_{random}$ & 0.5078 (0.0116) & 0.4974 (0.0173) & 0.5717 (0.0571) & 0.5435 (0.0644)\\
Shallow & 0.3704 (0.0123) & 0.3509 (0.0144) & 0.6041 (0.0253) & 0.5795 (0.0258)\\
Deep & 0.5219 (0.0182) & 0.5050 (0.0126) & 0.6119 (0.0213) & 0.5924 (0.0121)\\
Transformer & 0.5104 (0.0159) & 0.4999 (0.0127) & 0.6069 (0.0291) & 0.5931 (0.0211)\\
\mname & \textbf{0.5313} (0.0124) & \textbf{0.5244} (0.0142) & \textbf{0.6196} (0.0259) & \textbf{0.5992} (0.0223)\\
% \hline
\bottomrule
\end{tabular}
%}
\end{table*}
\subsection{Prediction Tasks}
\label{ssec:exp_task}
To evaluate the impact of jointly learning the encounter structure, we use prediction tasks based on a single encounter, rather than a sequence of encounters, which was the experiment setup in \cite{choi2018mime}.
However, \mname can be readily combined with a sequence aggregator such as RNN or 1-D CNN to handle a sequence of encounters, and derive patient representations for patient-level prediction tasks.
Specifically, we test the models on the following tasks.
Parentheses indicate which dataset is used for each task.
\begin{itemize}[leftmargin=5.5mm]
\item \textbf{Graph reconstruction (Synthetic)}: Given an encounter with $N$ features, we train models to learn $N$ feature embeddings $\Cb$, and predict if there is an edge between every pair of features, by performing an inner-product between each feature embedding pairs $\cb_i$ and $\cb_j$ (\textit{i.e.} $N^2$ binary predictions).
We do not use \textbf{Deep} baseline for this task, as we need individual embeddings for all features $c_i$'s.
\item \textbf{Diagnosis-Treatment classification (Synthetic)}: We assign labels to an encounter if there are specific diagnosis ($d_1$ and $d_2$) and treatment code ($m_1$) connections.
Specifically, we assign "1" if it contains a $d_1$-$m_1$ connection, and "2" if it contains a $d_2$-$m_1$ connection.
We intentionally made the task difficult so that the models cannot achieve a perfect score by just basing their prediction on whether $d_1$, $d_2$ and $m_1$ exist in an encounter.
The prevalence for both labels are approximately $3.3\%$.
Further details are provided in Appendix~\ref{appendix:dxtreatment}. This is a multi-label prediction task using the visit representation $\vb$.
%Further detail on the conditional probabilities of \textit{Diagnosis1}, \textit{Diagnosis2} and \textit{Treatment1} are provided in Appendix~\ref{appendix:dxtreatment}.
\item \textbf{Masked diagnosis code prediction (Synthetic, eICU)}: Given an encounter record, we mask a random diagnosis code $d_i$. We train models to learn the embedding of the masked code to predict its identity, \textit{i.e.} a multi-class prediction. For \textbf{Shallow} and \textbf{Deep}, we use the visit embedding $\vb$ as a proxy for the masked code representation.
The row and the column of the conditional probability matrix $\Pb$ that correspond to the masked diagnosis were also masked to zeroes.
\item \textbf{Readmission prediction (eICU)}: Given an encounter record, we train models to learn the visit embedding $\vb$ to predict whether the patient will be admitted to the ICU again during the same hospital stay, \textit{i.e.,} a binary prediction. The prevalence is approximately $17.2\%$.
\item \textbf{Mortality prediction (eICU)}: Given an encounter record, we train models to learn the visit embedding $\vb$ to predict patient death during the ICU admission, \textit{i.e.,} a binary prediction. The prevalence is approximately $7.3\%$.
%\item \textbf{Medications prediction}: Given an encounter record with all medication codes removed, we train models to learn the visit representation $\vb$ to predict all medication codes, \textit{i.e.,} a multi-label prediction with the output dimension the size of the medication vocabulary.
%\item \textbf{Future diagnosis prediction}: Given an outpatient encounter, we train models to learn $\vb$ to predict all diagnosis codes of the next encounter \textit{i.e.} a multi-label prediction with the output dimension the size of the diagnosis vocabulary.
\end{itemize}
For each task, data were randomly divided into train, validation, and test set in 8:1:1 ratio for 5 times, yielding 5 trained models, and we report the average performance.
Note that the conditional probability matrix $\Pb$ was calculated only with the training set.
Further training details and hyperparameter settings are described in Appendix~\ref{appendix:implementation}.

\begin{table*}[t]
\caption{KL divergence between the normalized true adjacency matrix and the attention map. We also show the entropy of the attention map to indicate the sparseness of the attention distribution. Parentheses denote standard deviations.}
\label{table:structure_eval}
\footnotesize
%\resizebox{\columnwidth}{!}{%
\centering
\begin{tabular}{lcccccc}
% \hline
\toprule
%Dataset & Model & \begin{tabular}{@{}c@{}} Validation\\Accuracy\end{tabular} & \begin{tabular}{@{}c@{}} Test\\Accuracy\end{tabular}\\

& \multicolumn{2}{c}{\textbf{Graph Reconstruction}} & \multicolumn{2}{c}{\textbf{Diagnosis-Treatment Classification}} & \multicolumn{2}{c}{\textbf{Masked Diagnosis Code Prediction}}\\
%\hline
%Model & \begin{tabular}{@{}c@{}} Validation\\Accuracy\end{tabular} & \begin{tabular}{@{}c@{}} Test\\Accuracy\end{tabular} & \begin{tabular}{@{}c@{}} Validation\\Accuracy\end{tabular} & \begin{tabular}{@{}c@{}} Test\\Accuracy\end{tabular}\\
Model & KL Divergence & Entropy & KL Divergence & Entropy & KL Divergence & Entropy\\
% \hline
\midrule
GCN$_P$ & 8.4844 (0.0140) & 1.5216 (0.0044) & 8.4844 (0.0140) & 1.5216 (0.0040) & 8.4844 (0.0140) & 1.5216 (0.0044)\\
Transformer & 19.6268 (2.9114) & 1.7798 (0.1411) & 14.3178 (0.2084) & 1.9281 (0.0368) & 15.1837 (0.8646) & 1.9941 (0.0522)\\
\mname & 7.6490 (0.0476) & 1.8302 (0.0135) & 8.0363 (0.0305) & 1.6003 (0.0244) & 8.9648 (0.1944) & 1.3305 (0.0889)\\
% \hline
\bottomrule
 %& GCN & 0.2828 & 0.3024\\
 %f& GCN$_P$ & 0.1908 & 0.2096\\
 %f& shallow & 0.3488 & 0.3417\\
%feICU & deep & 0.3549 & 0.3547\\
 %f& Transformer & 0.3647 & 0.3581\\
 %f& \mname & 0.2182 & 0.2328\\
%f\hline
\end{tabular}
%\vspace{-3mm}
%}
\end{table*}
\begin{figure*}[h]
\centering
\includegraphics[width=.85\textwidth]{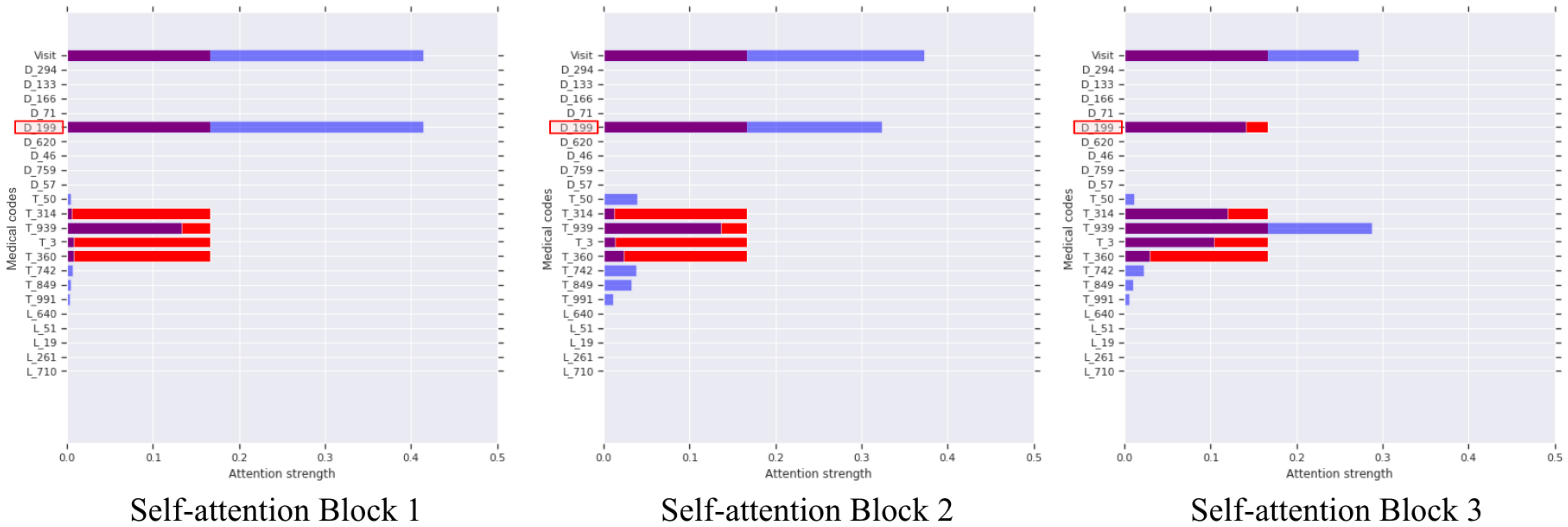}
\caption{
Attentions from each self-attention block of Transformer trained for graph reconstruction.
Code starting with `D' are diagnosis, `T' treatment, and `L' lab codes.
The diagnosis code with the red background D\_199 is attending to the other features.
The red bars indicate the codes that are actually connected to D\_199, and the blue bars indicate the attention given to each code.
}
\label{fig:recon_gct_att}
\end{figure*}
\subsection{Prediction Performance}
\label{ssec:exp_result}
%\input{tables/graph_recon_dt_pred_result.tex}
%\input{tables/graph_recon_dt_pred_result2.tex}
%\input{tables/graph_recon_result.tex}
%f\input{tables/dt_pred_result.tex}
%\input{tables/rx_pred_result.tex}
%\input{tables/dx_pred_result.tex}
Table~\ref{table:graph_recon_dt_pred_result} shows the graph reconstruction performance and the diagnosis-treatment classification performance of all models.
Naturally, GCN shows the best performance since it uses the true adajcency matrix $\Ab$. 
Given that GCN$_P$ is outperformed only by \mname, we can infer that the conditional probability is indeed indicative of the true structure.
\mname, which combines the strength of both GCN$_P$ and Transformer shows the best performance, besides GCN.
It is noteworthy that GCN$_{random}$ outperforms Shallow.
This seems to indicate that for graph reconstruction, attending to other features, regardless of how accurately the process follows the true structure, is better than individually embedding each feature.
Diagnosis-treatment classification, on the other hand, clearly penalizes randomly attending to the features, since GCN$_{random}$ shows the worst performance.
\mname again shows the best performance.

Table~\ref{table:node_pred_result} shows the model performance for masked diagnosis prediction for both datasets.
GCN could not be evaluated on eICU, since eICU does not have the true structure.
However, GCN naturally shows the best performance on the synthetic dataset.
%GCN$_{random}$ again demonstrates performance for both datasets, indicating that inferring a diagnosis based on other features requires accurate structure knowledge.
Interestingly, Transformer shows comparable performance to GCN$_{random}$, indicating the opposite nature of this task compared to graph reconstruction, where simply each feature attending to other features significantly improved performance.
Note that the task performance is significantly higher for eICU than for the synthetic dataset.
This is mainly due to eICU having a very skewed diagnosis code distribution.
In eICU, more than 80\% of encounters have diagnosis codes related to whether the patient has been in an operating room prior to the ICU admission. 
Therefore randomly masking one of them does not make the prediction task as difficult as for the synthetic dataset.
%Interestingly, both GCN$_P$ and \mname show strong performance on both datasets, but the performance gap is significantly wider for eICU.
%fThis seems due to the fact that eICU has a larger diagnosis code vocabulary than the synthetic dataset (2460 vs 1000), which makes guessing the correct diagnosis code harder.
%Also, eICU data volume is smaller than the synthetic dataset (33168 vs 50000), making the already harder task even more difficult for other models.
%Both GCN$_P$ and \mname, however, relies on the prior knowledge extracted from the training set, which is a form of rather strong regularization, making both models robust to the smaller dataset.

Table~\ref{table:mortality_readmission_result} shows the readmission prediction and mortality prediction performance of all models on eICU.
As shown by \mname's superior performance, it is evident that readmission prediction benefits from using the latent encounter structure.
Mortality prediction, on the other hand, seems to rely little on the encounter structure, as can be seen from the marginally superior performance of \mname compared to Transformer and Deep.
%demonstrate strikingly contrasting performance indicates that mortality prediction on eICU data relies little to the true structure.
%Rather, given the comparable performances of Shallow, it seems that mortality prediction can be successfully performed by simply adding all features.
%This seems natural, since eICU records are filtered for remote caregivers, who do not necessarily need to know information indicating patient death.
Even when the encounter structure seems unnecessary, however, \mname still outperforms all other models, demonstrating its potential to be used as a general-purpose EHR modeling algorithm.
%and it should be acting as a regularized Transformer, rather than trying to recover the true structure.
These two experiments indicate that not all prediction tasks require the true encounter structure, and it is our future work to apply \mname to various prediction tasks to evaluate its effectiveness.

%, Table~\ref{table:node_pred_result}, and Table~\ref{table:mortality_pred_result} respectively show mortality prediction performance, medication prediction performance, and masked diagnosis code prediction performance for all models on two datasets.

%fIn both mortality prediction and medication prediction, all models show stronger performance on MIMIC-III than eICU, which is reasonable given that MIMIC-III's encounter records are more dense, \textit{i.e.} there are more features per encounter on average as shown by Table~\ref{table:data_stat}.
%fIn masked code prediction, on the other hand, all models show stronger performance on eICU, naturally because eICU's diagnosis code vocabulary is smaller.
%fFurthermore, as MIMIC-III's encounter has more diagnosis codes on average, it is more likely for the models to predict one of the visible diagnosis codes.

%fAs all three tables show, Transformer outperforms baseline models on all \numtasks tasks on two different datasets.
%fThis empirical evidence strongly indicates Transformer's suitability to be used as a general-purpose EHR encounter modeling framework.
%fFinally, Transformer significantly outperforms baseline models on some tasks and datasets, while the perforamnce improvement is not as dramatic for other tasks and datasets, which requires further investigation in the future.
%fWe discuss Transformer's attention behavior in Appendix~\ref{appendix:attention}.

\subsection{Evaluating the Learned Encounter Structure}
\label{ssec:encounter_structure}
%Even though the ultimate goal of \mname is to outperform the \textit{bag-of-features} approach in prediction tasks on encounter records, the premise was that \mname is able to achieve this by effectively learning the encounter structure, even when the structure information is unavailable.
In this section, we analyze the learned structure of both Transformer and \mname.
As we know the true structure $\Ab$ of synthetic records, we can evaluate how well both models learned $\Ab$ via self-attention $\hat{\Ab}$.
Since we can view the normalized true adjacency matrix $\tilde{\Db}^{-1} \tilde{\Ab}$ as a probability distribution, we can measure how well the attention map $\hat{\Ab}$ in Eq.~\eqref{eq:define_adj} approximates $\tilde{\Db}^{-1} \tilde{\Ab}$ using KL divergence $D_{KL}(\tilde{\Db}^{-1} \tilde{\Ab} || \hat{\Ab})$.
Table~\ref{table:structure_eval} shows the KL divergence between the normalized true adjacency and the learned attention on the test set of the synthetic data while performing three different tasks.
For GCN$_P$, the adjacency matrix is fixed to the conditional probability matrix $\Pb$, so KL divergence can be readily calculated. 
For Transformer and \mname, we calculated KL divergence between $\tilde{\Db}^{-1} \tilde{\Ab}$ and the attention maps in each self-attention block, and averaged the results.
We repeated this process for 5 times (on 5 randomly sampled train, validation, test sets) and report the average performance.
Note that KL divergence can be lowered by evenly distributing the attention across all features, which is the opposite of learning the encounter structure.
Therefore we also show the entropy of $\hat{\Ab}$ alongside the KL divergence.

As shown by Table~\ref{table:structure_eval}, the conditional probabilities are closer to the true structure than what Transformer has learned, in all three tasks.
\mname shows similar performance to GCN$_P$ in all tasks, and was even able to improve upon $\Pb$ in both graph reconstruction and diagnosis-treatment classification tasks.
%Interestingly, it seems graph reconstruction can be done quite well even if the model deviates much from the true structure. Transformer's attention was significantly dissimilar to the true structure compared to GCN$_P$, but it was still able to outperform GCN$_P$.
It is notable that, despite having attentions significantly different from the true structure, Transformer demonstrated strong graph reconstruction performance in Table~\ref{table:graph_recon_dt_pred_result}.
This again indicates the importance of simply attending to other features in graph reconstruction, which was discussed in Section~\ref{ssec:exp_result} regarding the performance of GCN$_{random}$.
For the other two tasks, regularizing the models to stay close to $\Pb$ helped \mname outperform Transformer as well as other models.
%This seems to be due to the fact that graph reconstruction requires the features to ``see'' one another at least once to determine

\subsection{Attention Behavior Visualization}
In this section, we show the attention behavior of \mname when trained for graph reconstruction.
We randomly chose an encounter record from the test set of the synthetic dataset, which had less than 30 codes in order to enhance readability.
To show the attention distribution of a specific code, we chose the first diagnosis code connected to at least one treatment.
Figure~\ref{fig:recon_gct_att} shows \mname's attentions in each self-attention block when performing graph reconstruction.
Specifically we show the attention given by the diagnosis code D\_199 to other codes.
The red bars indicate the true connections, and the blue bars indicate the attention given to all codes.
%It can be seen that Transformer evenly attends to all codes in the first block, then develops its own attention.
%In the second block, it successfully recovers two of the true connections, but attends to incorrect codes in the third block.

Figure~\ref{fig:recon_gct_att} shows \mname's attention in each self-attention block when performing graph reconstruction.
As can be seen from the first self-attention block, \mname starts with a very specific attention distribution, as opposed to Transformer, which can be seen in Figure~\ref{supp_fig:recon_transformer_att} in Appendix~\ref{appendix:attention}.
The first two attentions given to the placeholder Visit node, and to itself are determined by the scalar value from Figure~\ref{fig:conditional_matrix}.
However, the attentions given to the treatment codes, especially T\_939 are derived from the conditional probability matrix $\Pb$.
Then in the following self-attention blocks, \mname starts to deviate from $\Pb$, and the attention distribution becomes more similar to the true adjacency matrix.
This nicely shows the benefit of using $\Pb$ as a guide to learning the encounter structure.
We further compare and analyze the attention behaviors of both Transformer and \mname under two different contexts, namely graph reconstruction and masked diagnosis code prediction, in Appendix~\ref{appendix:attention}.

\section{Conclusion}
\label{sec:conclusion}
Learning effective patterns from raw EHR data is an essential step for improving the performance of many downstream prediction tasks.
In this paper, we addressed the issue where the previous state-of-the-art method required the complete encounter structure information, and proposed \mname to capture the underlying encounter structure when the structure information is unknown. 
Experiments demonstrated that \mname outperformed various baseline models on encounter-based tasks on both synthetic data and a publicly available EHR dataset, demonstrating its potential to serve as a general-purpose EHR modeling algorithm.
%%The attention behavior indicated that Transformer is learning reasonable patterns for some features, but there is definitely room for improvement.
In the future, we plan to combine \mname with a sequence aggregator (\textit{e.g.} RNN) to perform patient-level prediction tasks such as heart failure diagnosis prediction or unplanned emergency admission prediction, while working on improving the attention mechanism to learn more medically meaningful patterns.

%\newpage
%\clearpage
\begin{quote}
\begin{small}
\bibliographystyle{aaai}
\bibliography{references}
\end{small}
\end{quote}

\newpage
\clearpage
\appendix
\begin{algorithm*}[h]
   \caption{Synthetic Encounter Records Generation}
   \label{alg:synthetic}
\begin{algorithmic}
   \STATE $|D| = 1000$ // Dx vocab size
   \STATE $|M| = 1000$ // Treatment vocab size
   \STATE $|R| = 1000$ // Lab vocab size
   \STATE 
   \STATE \textit{// Independent diagnosis occurrence probability}
   \STATE $p(D)$ = permute(normalize(numpy.random.pareto($\alpha=2.0$, size=|D|)))
   \STATE 
   \STATE \textit{// Conditional probability of a diagnosis co-occurring with another diagnosis}
   \STATE $p(D|d_1)$ = permute(normalize(numpy.random.pareto($\alpha=1.5$, size=|D|)))
   \STATE ...
   \STATE $p(D|d_{|D|})$ = permute(normalize(numpy.random.pareto($\alpha=1.5$, size=|D|)))
   \STATE 
   \STATE \textit{// Conditional probability of a treatment being ordered for a specific diagnosis}
   \STATE $p(M|d_1)$ = permute(normalize(numpy.random.pareto($\alpha=1.5$, size=|M|)))
   \STATE ...
   \STATE $p(M|d_{|D|})$ = permute(normalize(numpy.random.pareto($\alpha=1.5$, size=|M|)))
   \STATE 
   \STATE \textit{// Conditional probability of a lab being ordered for a specific treatment and diagnosis}
   \STATE $p(R|m_1, d_1)$ = permute(normalize(numpy.random.pareto($\alpha=1.5$, size=|R|)))
   \STATE ...
   \STATE $p(R|m_{|M|}, d_{|D|})$ = permute(normalize(numpy.random.pareto($\alpha=1.5$, size=|R|)))
   \STATE 
   \STATE \textit{// Bernoulli probability to determine whether to sample a diagnosis code given a previous diagnosis. Values are clipped to [0., 1.).}
   \STATE $a(D)$ = numpy.random.normal($\mu=0.5$, $\sigma=0.1$, size=|D|))
   \STATE 
   \STATE \textit{// Bernoulli probability to determine whether to sample a treatment code given a diagnosis. Values are clipped to [0., 1.).}
   \STATE $b(D)$ = numpy.random.normal($\mu=0.5$, $\sigma=0.25$, size=|M|))
   \STATE 
   \STATE \textit{// Bernoulli probability to determine whether to sample a lab code given a treatment and a diagnosis. Values are clipped to [0., 1.).}
   \STATE $c(M, D)$ = numpy.random.normal($\mu=0.5$, $\sigma=0.25$, size=(|M|, |D|))
   \STATE
   \STATE \textit{// Start creating synthetic records}
   \REPEAT %// Sample independent diagnosis codes
   \STATE Sample a diagnosis $d_i$ from $p(D)$
   \REPEAT %// Sample diagnosis conditioned on previous diagnosis
   \STATE Sample a diagnosis $d_j$ from $p(D|d_j)$
   \UNTIL{$x \sim Uniform(0, 1) < a(d_i)$}
   \UNTIL{$x \sim Uniform(0, 1) < 0.5$}
   \FOR{$d_i$ in the sampled diagnosis codes}
   \REPEAT
   \STATE Sample a treatment $m_j$ from $P(M|d_i)$.
   \REPEAT
   \STATE Sample a lab $r_k$ from $P(R|m_j,d_i)$.
   \UNTIL{$x \sim Uniform(0, 1) < c(m_j, d_i)$}
   \UNTIL{$x \sim Uniform(0, 1) < b(d_i)$}
   \ENDFOR
\end{algorithmic}
\end{algorithm*}
\section{Generating Synthetic Encounter Records}
\label{appendix:synthetic}
Due to the large volume of the synthetic records, we cannot submit the data as supplementary material.
Also as per the ``no link in the supplementary material'' policy, we do not provide a link to download the synthetic records.
We will open-source the synthetic records along with the codes in the future.
Instead, we describe the synthetic data creation process in this section.

As described in Section~\ref{ssec:exp_synthetic}, we use the Pareto distribution to capture the long-tailed nature of medical codes.
We also define $a(D), b(D)$, and $c(M, D)$ to determine when to stop sampling the codes.
The overall generation process starts by sampling a diagnosis code. Then we sample a diagnosis code that is likely to co-occur with the previous sampled diagnosis code.
After the diagnosis codes are sampled, we iterate through the sampled diagnosis code to sample a treatment code that is likely to be ordered for each diagnosis code. At the same time as sampling the treatment code, we sample lab codes that are likely to be produced by each treatment code.
The overall algorithm is described in Algorithm~\ref{alg:synthetic}.

Note that we use $p(M|D)$ to model the treatment being ordered due to a diagnosis code, instead of $P(M|M,D)$, which might be more accurate since a treatment may depend on the already ordered treatments as well. However, we assume that given a diagnosis code, treatments that follow are conditionally independent, therefore each treatment can be factorized by $p(M|D)$. The same assumption went into using $P(R|M,D)$, instead of $P(R|R,M,D)$.

Finally, among the generated synthetic encounters, we removed the ones that had less than 5 diagnosis or treatment codes, in order to make the encounter structure sufficiently complex.
Additionally, we removed encounters which contained more than 50 diagnosis or treatment or lab codes in order to make the encounter structure realistic (\textit{i.e.} it is unlikely that a patient receives more than 50 diagnosis codes in one hospital encounter). 
For the eICU dataset, we also removed the encounters with more than 50 diagnosis or treatment codes.
But we did not remove any encounters for having less than 5 diagnosis or treatment codes, as that would leave us only approximately 7,000 encounter records, which are rather small for training neural networks.

\section{Diagnosis-Treatment Classification Task}
\label{appendix:dxtreatment}
This task is used to test the model's ability to derive a visit representation $\vb$ (\textit{i.e.} graph-level representation) that correctly preserves the encounter structure.
As described in Section~\ref{ssec:exp_task}, this is a multi-label classification problem, where an encounter is assigned the label ``1'' if it contains a connected pair of a diagnosis code $d_1$ and a treatment code $m_1$ (\textit{i.e.} $m_1$ was ordered because of $d_1$). An encounter is assigned the label ``2'' if it contains a connected pair of $d_2$ and $m_1$. Therefore it is possible that an encounter is assigned both labels ``1'' and ``2'', or not assigned any label at all.

Since we want to test the model's ability to correctly learn the encounter structure, we do not want the model to achieve a perfect score, for example, by just predicting label ``1'' based on whether both $d_1$ and $m_1$ simply exist in an encounter.
Therefore we adjusted the sampling probabilities to make this task difficult. Specifically, we set $p(d_1)=0.33, a(d_1)=0.8, p(d_2 | d_1)=0.33, b(d_1)=0.5, b(d_2)=0.5, (p(m_1 | d_1)=0.2, p(m_1 | d_2)=0.8$.
Therefore the probability of an encounter containing a $d_1$-$m_1$ connection is $p(d_1)b(d_1)p(m_1 | d_1) = 0.33 \times 0.5 \times 0.2 = 0.033$.
The probability of an encounter contaning a $d_2$-$m_1$ connection is $p(d_1)a(d_1)p(d_2|d_1)b(d_2)p(m_1 | d_2) = 0.33 \times 0.8 \times 0.33 \times 0.5 \times 0.8 \approx 0.033$.
Therefore The overall probability of the two connection pairs occurring in an encounter are more or less the same, and the model cannot achieve a perfect score unless the model correctly identifies the encounter structure.

\section{Training Details}
\label{appendix:implementation}
All models were trained with Adam~\cite{kingma2014adam} on the training set, and performance was evaluated against the validation set to select the final model.
Final performance was evaluated against the test set.
We used the minibatch of size 32, and trained all models for 1,000,000 iterations (\textit{i.e.} minibatch updates), which was sufficient for convergence for all tasks.
After an initial round of preliminary experiments, the embedding size of the encounter features was set to 128.
For GCN, GCN$_P$, GCN$_{random}$, Transformer, and \mname, we used undirected adjacency/attention matrix to enhance the message passing efficiency.
All models were implemented in TensorFlow 1.13 \cite{abadi2016tensorflow}, and trained with a system equipped Nvidia P100's.

Tunable hyperparameters for models \textbf{Shallow}, \textbf{Deep}, \textbf{GCN}, \textbf{GCN$_P$}, \textbf{GCN$_{random}$}, and \textbf{Transformer} are as follows:
\begin{itemize}[leftmargin=5.5mm]
    \item Adam learning rate ($0.0001 \sim 0.1$)
    \item Drop-out rate between layers ($0.0 \sim 0.9$)
\end{itemize}
Transformer used three self-attention blocks, which was sufficient to cover the entire depth of EHR encounters.
Shallow used 15 feedforward layers and Deep used 8 feedforward layers before, and 7 feedforward layers after summing the embeddings.
The number of layers were chosen to match the number of trainable parameters of Transformer and \mname.
GCN, GCN$_P$ and GCN$_{random}$ used 5 convolution steps to match the number of trainable parameters of Transformer.
Transformer used one attention head to match its representative power to GCN, GCN$_P$, and GCN$_{random}$, and so that we can accurately evaluate the effect of learning the correct encounter structure.

Tunable hyperparameters for \textbf{GCT} are as follows:
\begin{itemize}[leftmargin=5.5mm]
    \item Adam learning rate ($0.0001 \sim 0.1$)
    \item Drop-out rate between layers ($0.0 \sim 0.9$)
    \item Regularization coefficient ($0.01 \sim 100.0$)
\end{itemize}
\mname also used three self-attention blocks and one attention head.
All Hyperparameters were searched via bayesian optimization with Gaussian Process for 72-hour wall clock time based on one of the five randomly sampled train/validation/test set.
Then the chosen hyperparameters were used for training models on all five sets.
Hyperparameters used for each task is described below in Table~\ref{table:hyperparams}.
%\ref{table:hyper_graph_recon}, \ref{table:hyper_dt_classification}, \ref{table:hyper_node_pred_synthetic}, \ref{table:hyper_node_pred_eicu}, \ref{table:hyper_readmission}, and \ref{table:hyper_mortality}.
\begin{table*}[h]
\caption{Hyperparameters for all tasks.}
\vspace{1mm}
\label{table:hyperparams}
\footnotesize
\centering
\begin{tabular}{lccccccc}
% \hline
\toprule
% \hline
\multicolumn{8}{c}{Hyperparameters for graph reconstruction on the synthetic data.}\\
& GCN & GCN$_P$ & GCN$_{random}$ & Shallow & Deep & Transformer & \mname\\
\midrule
Learning rate & 0.00045 & 0.0006 & 0.0003 & 0.00025 & - & 0.0007 & 0.0005\\
MLP dropout rate & 0.3 & 0.01 & 0.5 & 0.2 & - & 0.8 & 0.3\\
Post-MLP dropout rate & 0.2 & 0.02 & 0.005 & - & - & 0.001 & 0.1\\
Regularization coef. & - & - & - & - & - & - & 0.02\\
\midrule
\multicolumn{8}{c}{Hyperparameters for diagnosis-treatment classification on the synthetic data.}\\
& GCN & GCN$_P$ & GCN$_{random}$ & Shallow & Deep & Transformer & \mname\\
\midrule
Learning rate & 0.0001 & 0.0001 & 0.0001 & 0.0002 & 0.0008 & 0.00015 & 0.0001\\
MLP dropout rate & 0.2 & 0.3 & 0.5 & 0.02 & 0.01 & 0.5 & 0.85\\
Post-MLP dropout rate & 0.65 & 0.02 & 0.4 & - & 0.3 & 0.01 & 0.03\\
Regularization coef. & - & - & - & - & - & - & 0.05\\
\midrule
\multicolumn{8}{c}{Hyperparameters for masked diagnosis code prediction on the synthetic data.}\\
& GCN & GCN$_P$ & GCN$_{random}$ & Shallow & Deep & Transformer & \mname\\
\midrule
Learning rate & 0.0003 & 0.0007 & 0.0002 & 0.0007 & 0.0004 & 0.0003 & 0.0001\\
MLP dropout rate & 0.01 & 0.8 & 0.5 & 0.08 & 0.12 & 0.4 & 0.85\\
Post-MLP dropout rate & 0.88 & 0.005 & 0.5 & - & 0.75 & 0.5 & 0.6\\
Regularization coef. & - & - & - & - & - & - & 0.05\\
\midrule
\multicolumn{8}{c}{Hyperparameters for masked diagnosis code prediction on eICU.}\\
& GCN & GCN$_P$ & GCN$_{random}$ & Shallow & Deep & Transformer & \mname\\
\midrule
Learning rate & - & 0.0005 & 0.0001 & 0.0001 & 0.00012 & 0.0001 & 0.0009\\
MLP dropout rate & - & 0.5 & 0.3 & 0.3 & 0.4 & 0.87 & 0.5\\
Post-MLP dropout rate & - & 0.5 & 0.4 & - & 0.45 & 0.2 & 0.03\\
Regularization coef. & - & - & - & - & - & - & 50.0\\
\midrule
\multicolumn{8}{c}{Hyperparameters for readmission prediction on eICU.}\\
& GCN & GCN$_P$ & GCN$_{random}$ & Shallow & Deep & Transformer & \mname\\
\midrule
Learning rate & - & 0.00024 & 0.0001 & 0.0001 & 0.00011 & 0.0002 & 0.00022\\
MLP dropout rate & - & 0.3 & 0.7 & 0.63 & 0.05 & 0.45 & 0.08\\
Post-MLP dropout rate & - & 0.1 & 0.01 & - & 0.33 & 0.28 & 0.024\\
Regularization coef. & - & - & - & - & - & - & 0.1\\
\midrule
\multicolumn{8}{c}{Hyperparameters for mortality prediction on eICU.}\\
& GCN & GCN$_P$ & GCN$_{random}$ & Shallow & Deep & Transformer & \mname\\
\midrule
Learning rate & - & 0.0003 & 0.00013 & 0.0001 & 0.00015 & 0.0006 & 0.00011\\
MLP dropout rate & - & 0.85 & 0.9 & 0.25 & 0.01 & 0.88 & 0.72\\
Post-MLP dropout rate & - & 0.04 & 0.01 & - & 0.01 & 0.2 & 0.005\\
Regularization coef. & - & - & - & - & - & - & 1.5\\
\bottomrule
\end{tabular}
\end{table*}

\section{Prediction Performance in AUROC}
\label{appendix:auroc}
\begin{table*}[h]
\caption{Graph reconstruction and diagnosis-treatment classification performance measuerd in AUROC. Parentheses denote standard deviations.}
\label{table:graph_recon_dt_pred_auroc}
\footnotesize
%\resizebox{\columnwidth}{!}{%
\centering
\begin{tabular}{lcccc}
% \hline
\toprule
& \multicolumn{2}{c}{\textbf{Graph reconstruction}} & \multicolumn{2}{c}{\textbf{Diagnosis-Treatment classification}}\\
%\hline
Model & Validation AUROC & Test AUROC & Validation AUROC & Test AUROC\\
%Model & \begin{tabular}{@{}c@{}} Validation\\AUROC\end{tabular} & \begin{tabular}{@{}c@{}} Test\\AUROC\end{tabular} & \begin{tabular}{@{}c@{}} Validation\\AUROC\end{tabular} & \begin{tabular}{@{}c@{}} Test\\AUROC\end{tabular}\\
% \hline
\midrule
GCN & 1.0 (0.0) & 1.0 (0.0) & 1.0 (0.0) & 1.0 (0.0)\\
% \hline
\midrule
GCN$_P$ & 0.8870 (0.0011) & 0.8865 (0.0005) & 0.9493 (0.0127) & 0.9475 (0.0135)\\
GCN$_{random}$ & 0.8806 (0.0009) & 0.8799 (0.0008) & 0.9230 (0.0053) & 0.9221 (0.0070)\\
Shallow & 0.8578 (0.0010) & 0.8573 (0.0005) & 0.9575 (0.0116) & 0.9584 (0.0140)\\
Deep & - & - &  0.9387 (0.0071) & 0.9374 (0.0041)\\
Transformer & 0.8843 (0.0013) & 0.8844 (0.0008) & 0.9494 (0.0226) & 0.9493 (0.0210)\\
\mname & \textbf{0.8936} (0.0012) & \textbf{0.8931} (0.0013) & \textbf{0.9626} (0.0146) & \textbf{0.9600} (0.0154)\\
% \hline
\bottomrule
\end{tabular}
%f}
\end{table*}
\begin{table*}[h]
\caption{Readmission prediction and mortality prediction performance on eICU measured in AUROC. Parentheses denote standard deviations.}
\label{table:mortality_readmission_auroc}
\footnotesize
%\resizebox{\columnwidth}{!}{%
\centering
\begin{tabular}{lcccc}
% \hline
\toprule
& \multicolumn{2}{c}{\textbf{Readmission prediction}} & \multicolumn{2}{c}{\textbf{Mortality prediction}}\\
%\hline
Model & Validation AUROC & Test AUROC & Validation AUROC & Test AUROC\\
%Model & \begin{tabular}{@{}c@{}} Validation\\AUROC\end{tabular} & \begin{tabular}{@{}c@{}} Test\\AUROC\end{tabular} & \begin{tabular}{@{}c@{}} Validation\\AUROC\end{tabular} & \begin{tabular}{@{}c@{}} Test\\AUROC\end{tabular}\\
% \hline
\midrule
GCN$_P$ & 0.7403 (0.0078) & 0.7355 (0.0081) & 0.8971 (0.0047) & 0.8953 (0.0065)\\
GCN$_{random}$ & 0.7243 (0.0046) & 0.7259 (0.0080) & 0.8939 (0.0243) & 0.8941 (0.0220)\\
Shallow & 0.6794 (0.0129) & 0.6734 (0.0101) & 0.9000 (0.0083) & 0.8972 (0.0038)\\
Deep & 0.7478 (0.0124) & 0.7412 (0.0074) & \textbf{0.9101} (0.0057) & 0.9092 (0.0060)\\
Transformer & 0.7333 (0.0065) & 0.7301 (0.0101) & 0.9089 (0.0121) & 0.9017 (0.0152)\\
\mname & \textbf{0.7525} (0.0128) & \textbf{0.7502} (0.0114) & 0.9089 (0.0052) & \textbf{0.9120} (0.0048)\\
% \hline
\bottomrule
\end{tabular}
%}
\end{table*}
Table \ref{table:graph_recon_dt_pred_auroc} shows the graph reconstruction performance and the diagnosis-treatment classification performance of all models measured in AUROC.
Table \ref{table:mortality_readmission_auroc} shows the readmission prediction performance and the mortality prediction performance of all models measured in AUROC.
We can readily see that \mname outperforms all other models in all tasks in terms of AUROC as well.

\section{Attention Behavior}
\label{appendix:attention}
\begin{figure*}[h]
\centering
\includegraphics[width=.95\textwidth]{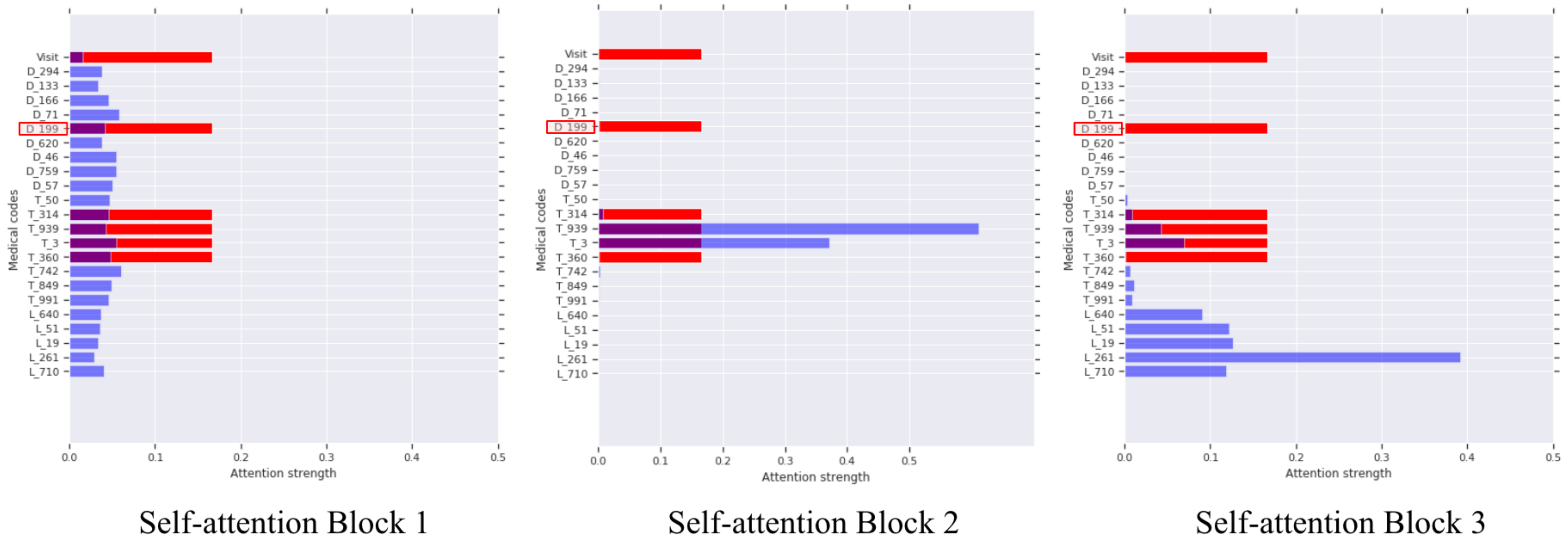}
\caption{
Attentions from each self-attention block of Transformer trained for graph reconstruction.
Code starting with `D' are diagnosis codes, `T' treatment codes, `L' lab codes.
The diagnosis code with the red background D\_199 is attending to the other features.
The red bars indicate the codes that are actually connected to D\_199, and the blue bars indicate the attention given to all codes.
}
\label{supp_fig:recon_transformer_att}
\end{figure*}

\begin{figure*}[h]
\centering
\includegraphics[width=.95\textwidth]{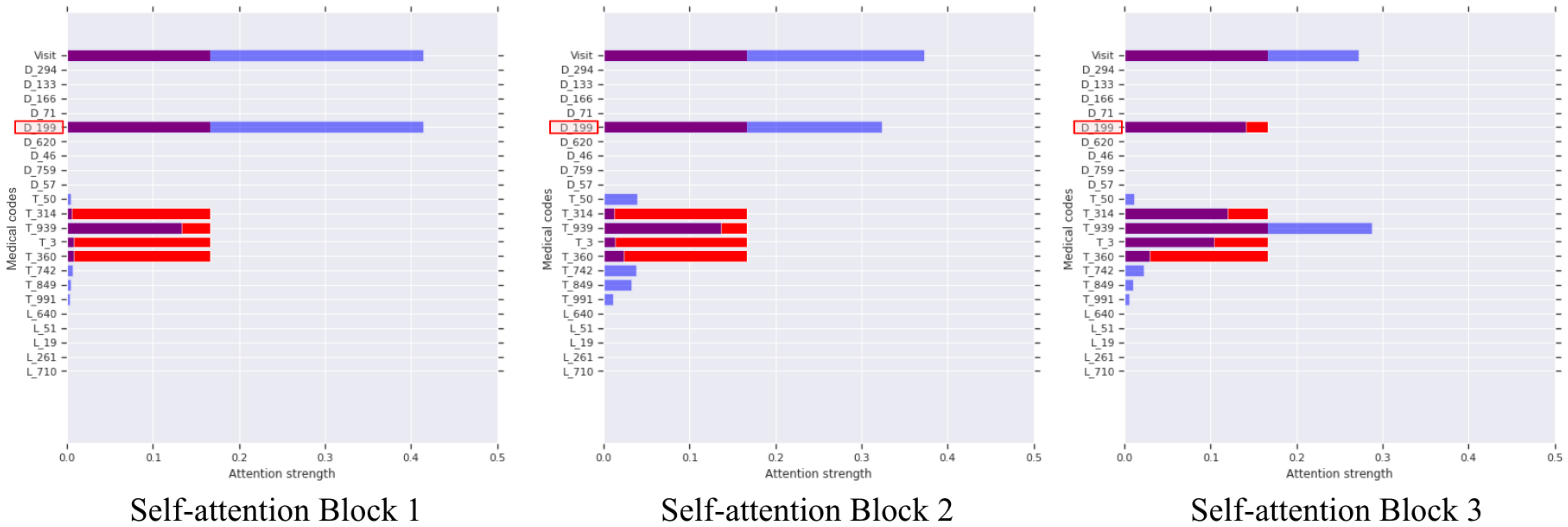}
\caption{
Attentions from each self-attention block of \mname trained for graph reconstruction.
Code starting with `D' are diagnosis codes, `T' treatment codes, `L' lab codes.
The diagnosis code with the red background D\_199 is attending to the other features.
The red bars indicate the codes that are actually connected to D\_199, and the blue bars indicate the attention given to all codes.
}
\label{supp_fig:recon_gct_att}
\end{figure*}

\begin{figure*}[h]
\centering
\includegraphics[width=.95\textwidth]{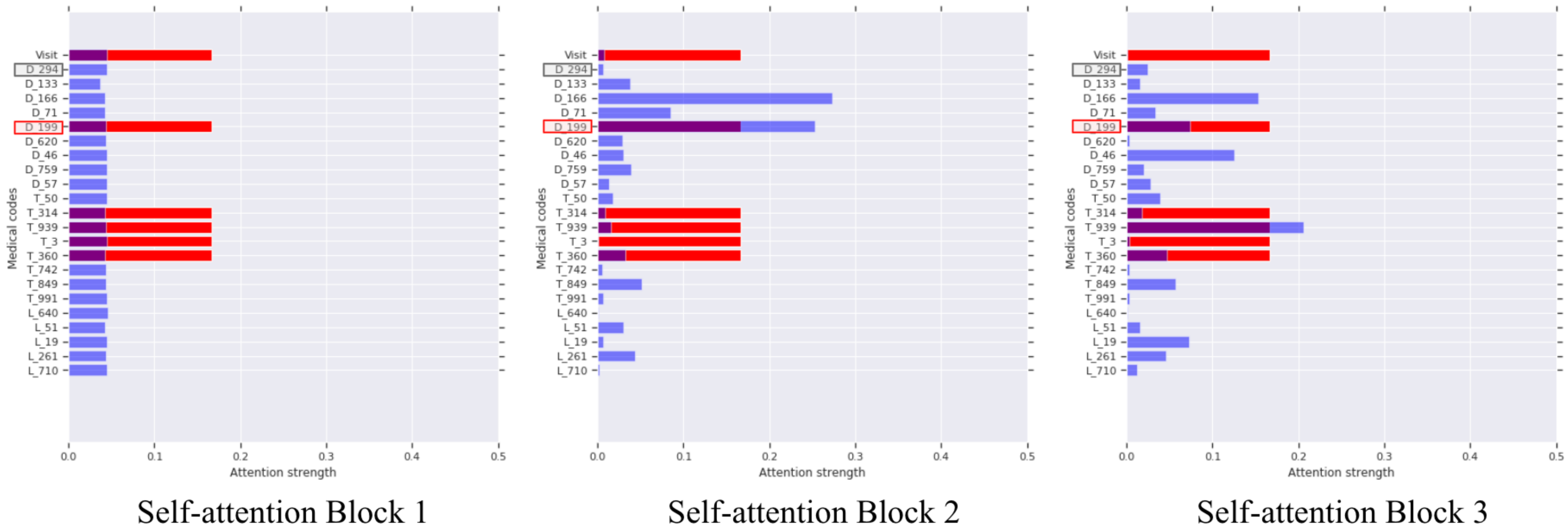}
\caption{
Attentions from each self-attention block of Transformer trained for masked diagnosis code prediction.
Code starting with `D' are diagnosis codes, `T' treatment codes, `L' lab codes.
The diagnosis code with the red background D\_199 is attending to the other features.
The diagnosis code with the gray background D\_294 is the masked diagnosis code.
The red bars indicate the codes that are actually connected to D\_199, and the blue bars indicate the attention given to all codes.
}
\label{supp_fig:node_transformer_att}
\end{figure*}

\begin{figure*}[h]
\centering
\includegraphics[width=.95\textwidth]{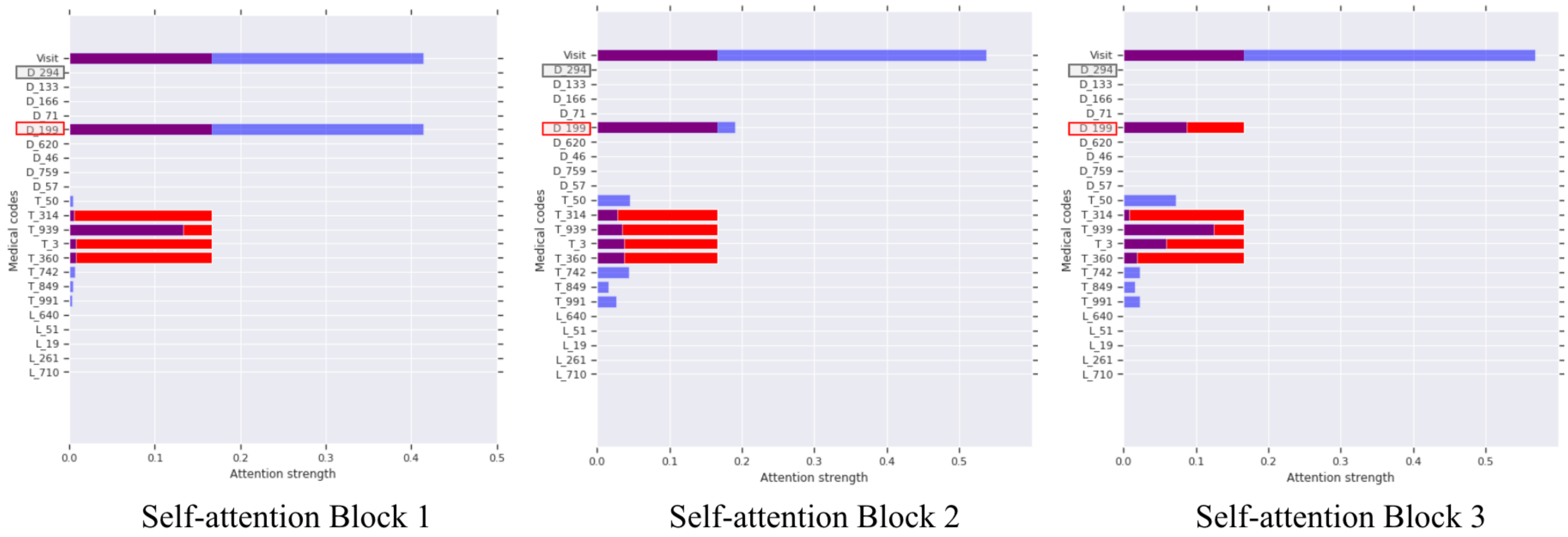}
\caption{
Attentions from each self-attention block of \mname trained for masked diagnosis code prediction.
Code starting with `D' are diagnosis codes, `T' treatment codes, `L' lab codes.
The diagnosis code with the red background D\_199 is attending to the other features.
The diagnosis code with the gray background D\_294 is the masked diagnosis code.
The red bars indicate the codes that are actually connected to D\_199, and the blue bars indicate the attention given to all codes.
}
\label{supp_fig:node_gct_att}
\end{figure*}

In this section, we compare the attention behavior of Transformer and \mname in two different context; graph reconstruction and masked diagnosis code prediction.
We randomly chose an encounter record from the test set of the synthetic dataset, which had less than 30 codes in order to enhance readability.
To show the attention distribution of a specific code, we chose the first diagnosis code connected to at least one treatment.
Figure~\ref{supp_fig:recon_transformer_att} shows Transformer's attentions in each self-attention block when performing graph reconstruction.
Specifically we show the attention given by the diagnosis code D\_199 to other codes.
The red bars indicate the true connections, and the blue bars indicate the attention given to all codes.
It can be seen that Transformer evenly attends to all codes in the first block, then develops its own attention.
In the second block, it successfully recovers two of the true connections, but attends to incorrect codes in the third block.

Figure~\ref{supp_fig:recon_gct_att} shows \mname's attention in each self-attention block when performing graph reconstruction.
Contrary to Transformer, \mname starts with a very specific attention distribution.
The first two attentions given to the placeholder Visit node, and to itself are determined by the scalar value from Figure~\ref{fig:conditional_matrix}.
However, the attentions given to the treatment codes, especially T\_939 are derived from the conditional probability matrix $\Pb$.
Then in the following self-attention blocks, \mname starts to deviate from $\Pb$, and the attention distribution becomes more similar to the true adjacency matrix.
This nicely shows the benefit of using $\Pb$ as a guide to learning the encounter structure.

Since the goal of the graph reconstruction task is to predict the edges between nodes, it may be an obvious result that both Transformer and \mname's attentions mimic the true adjacency matrix.
Therefore, we show another set of attentions from Transformer and \mname trained for the masked diagnosis code prediction task.
Figure~\ref{supp_fig:node_transformer_att} shows Transformer's attention while performing the masked diagnosis code prediction.
Note that the diagnosis code D\_294 is masked, and therefore the model does not know its identity.
Similar to graph reconstruction, Transformer starts with an evenly distributed attentions, and develops its own structure.
Interestingly, it learns to attend to the right treatment in the third block, but mostly tries to predict the masked node's identity by attending to other diagnosis codes, while mostly ignoring the lab codes.

Figure~\ref{supp_fig:node_gct_att} shows \mname's attention while performing the masked diagnosis code prediction task.
Again, \mname starts with the conditional probability matrix $\Pb$, then develops its own attention.
But this time, understandably, the attention maps are not as similar to the true structure as in the graph reconstruction task.
An interesting finding is that \mname attends heavily to the placeholder Visit node in this task.
This is inevitable, given that we only allow diagnosis codes to attend to treatment codes (see the white cells in Figure~\ref{fig:conditional_matrix}), and therefore, if \mname wants to look at other diagnosis codes, it can only be done by indirectly receiving information via the Visit node.
And as Figure~\ref{supp_fig:node_transformer_att} suggests, predicting the identity of the masked code seems to require knowing the co-occurring diagnosis codes as well as the treatment codes.
Therefore, unlike in the graph reconstruction task, \mname puts heavy attention to the Visit node in this task, in order to learn the co-occurring diagnosis codes.

\end{document}